\documentclass[letterpaper]{article} 
\usepackage{aaai24}  
\usepackage{times}  
\usepackage{helvet}  
\usepackage{courier}  
\usepackage[hyphens]{url}  
\usepackage{graphicx} 
\usepackage{xcolor} 
\usepackage{natbib}  
\usepackage{caption} 
\usepackage{algorithm}
\usepackage{algorithmic}

\usepackage{newfloat}
\usepackage{listings}
\DeclareCaptionStyle{ruled}{labelfont=normalfont,labelsep=colon,strut=off} 
\floatstyle{ruled}
\newfloat{listing}{tb}{lst}{}
\floatname{listing}{Listing}
\usepackage{amsmath}
\usepackage{amssymb}
\usepackage{mathtools}

\DeclareMathOperator*{\argmin}{argmin}
\DeclareMathOperator*{\sign}{sign}
\DeclareMathOperator*{\argmax}{argmax}

\title{Adversarial Training Using Feedback Loops}
\author{
    Ali Haisam Muhammad Rafid\textsuperscript{\rm 1},
    Adrian Sandu\textsuperscript{\rm 1}
}
\affiliations{
    \textsuperscript{\rm 1}Virginia Tech, Blacksburg, Virginia-24060, USA\\
    haisamrafid@vt.edu, sandu@cs.vt.edu
    
}

\begin{document}

\onecolumn

\thispagestyle{empty}
\setcounter{page}{0}

\makeatletter
\def\Year#1{%
  \def\yy@##1##2##3##4;{##3##4}%
  \expandafter\yy@#1;
}
\makeatother

\begin{Huge}
\begin{center}
Computational Science Laboratory Technical Report CSL-TR-23-5 \\
August 22, 2023
\end{center}
\end{Huge}
\vfil
\begin{huge}
\begin{center}
Ali Haisam Muhammad Rafid\\
Adrian Sandu
\end{center}
\end{huge}

\vfil
\begin{huge}
\begin{it}
\begin{center}
``Adversarial Training Using Feedback Loops''
\end{center}
\end{it}
\end{huge}
\vfil

\begin{large}
\begin{center}
Computational Science Laboratory \\
Computer Science Department \\
Virginia Polytechnic Institute and State University \\
Blacksburg, VA 24060 \\
Phone: (540)-231-2193 \\
Fax: (540)-231-6075 \\ 
Email: \url{sandu@cs.vt.edu} \\
Web: \url{http://csl.cs.vt.edu}
\end{center}
\end{large}

\vspace*{1cm}

\begin{tabular}{ccc}
\includegraphics[width=2.5in]{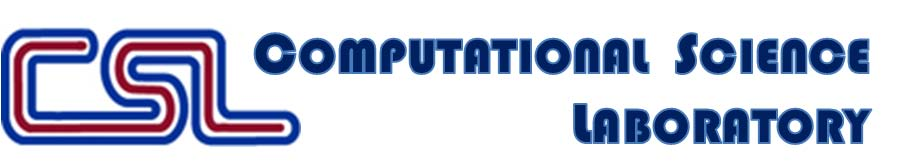}
&\hspace{2.5in}&
\includegraphics[width=2.5in]{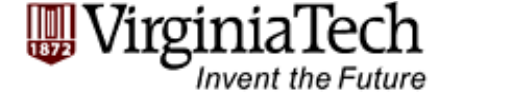} \\
{\bf\em\large Compute the Future} &&\\
\end{tabular}
\newpage

\twocolumn

\maketitle

\begin{abstract}
Deep neural networks (DNN) have found wide applicability in numerous fields due to their ability to accurately learn very complex input-output relations. Despite their accuracy and extensive use, DNNs are highly susceptible to adversarial attacks due to limited generalizability. For future progress in the field, it is essential to build DNNs that are robust to \textit{any} kind of perturbations to the data points. In the past, many techniques have been proposed to robustify DNNs using first-order derivative information of the network. 

This paper proposes a new robustification approach based on control theory. A neural network architecture that incorporates feedback control, named Feedback Neural Networks, is proposed. The controller is itself a neural network, which is trained using regular and adversarial data such as to stabilize the system outputs. {\it The novel adversarial training approach based on the feedback control architecture is called Feedback Looped Adversarial Training (FLAT).} Numerical results on standard test problems empirically show that our FLAT method is more effective than the state-of-the-art to guard against adversarial attacks.
\end{abstract}

\section{Introduction}
\label{sec:intro}

Deep neural networks are ubiquitous and can be found in countless applications today. While their predictive powers are remarkable, trained neural networks are prone to adversarial attacks. Specifically, a malicious player can perturb a regular input (which the trained network maps to a correct result) with a small, but well-chosen perturbation, such that the network maps the perturbed input to an incorrect output.  The goal of this work is to develop a methodology for constructing \textit{robust neural networks}, where small changes in the input do not lead to large changes in the output.

This issue of adversarial attacks has been widely studied with many different attack methods created \cite{Bai_Zeng_Jiang_Xia_Ma_Wang_2022}. The training procedure colloquially referred to as \textit{``adversarial training''} uses such maliciously perturbed inputs paired with the correct outputs to enrich the training set, and to robustify the neural networks model against such attacks. In this approach, guarding against adversarial attacks starts with specifying the type and method of attack itself. The \textit{attack model} limits the perturbation in the input data according to a prescribed criterion, such that the benign and malign examples are not human-distinguishable. For example, adversarial examples are limited to live in a small-radius $\ell_\infty$ ball around the benign example. 

Fast gradient sign method (FGSM) is perhaps the best known adversarial attack. First introduced by \cite{goodfellow2014explaining}, this one-step method perturbs the input along the direction of the fastest increase in the value of the loss function. Later improvements added multiple  steps, clipping the gradients, and projecting the perturbed input onto a desired attack model \cite{Kurakin2016Jul}. The gradient-based update has also been equipped with a memory term to create a momentum-based attack \cite{Dong2017Oct}. In \cite{carlini2017towards} it is proposed to perturb an input based on a distance metric until the input is classified as a different target class; three attack methods are proposed, based on three different distance metrics. In \cite{madry2017towards}, the authors proposed the PGD (projected gradient descent) attack, where a perturbation of some step size is projected onto a ball of interest of allowed maximum perturbation. The  PGD attack also has a parameter-free extension described in \cite{croce2020reliable}.

To fend off different adversarial attacks, various defense methods have been developed in the literature. \cite{zhang2019theoretically} proposes a new loss function named TRADES, which uses the loss between the natural image and the perturbed image as the regularizer to the original loss function, while also adversarially training a neural network. In \cite{Wang2020Improving}, the loss function is based on whether a perturbed input is misclassified or not. Friendly adversarial training (FAT) proposed in \cite{zhang2020attacks}, employs early-stopped PGD so that only the adversarial examples that help with minimizing the loss function are used. The training algorithm described in \cite{zhang2020geometry} takes into consideration the distance of a training example to the decision boundary; the idea is that examples closer to the decision boundary can be more easily misclassified, and hence should be prioritized over examples that are further away from the boundary. A recent adversarial training method proposed in \cite{wang2022self} utilizes an ensemble based training approach to adversarially train a model without training an ensemble of classifiers. Instead of using a set of classifiers, the authors compute the optimal weights of the model by considering the trajectory of the weights as an exponential moving average. This considers current weights and also has influence from previous weights.

Negative feedback control is a very general mechanism by which the outputs of a system are stabilized in the presence of unknown perturbations \cite{lewis2012optimal}.  Specifically, in a feedback control system a controller monitors the values of output variables, and compares them with the reference values. The difference between the actual and desired values is fed back, and is used to generate a control action that nudges the system output back toward the reference values. Therefore a control system is able to self-correct, and maintain a stable output, regardless of the disturbances that push the output away from reference values. 

As a simple example of negative feedback, consider a sink into which water flows from an open faucet; the faucet inflow rate is under our control.  In the same time, the water flows out of the sink through a drain; the outflow rate is variable, and is under the control of an adversary. Our goal is to keep the water level at a given constant height. However, for any inflow level we set, the adversary can change the drain outflow such as to prevent the water stabilizing at the desired height. A control system measures the difference between the current and desired water levels; if this difference is negative (the current level is lower) then the control action is to turn the faucet up such as to increase the inflow; vice-versa, if the difference is positive (the current level is higher) then the control action is to turn the faucet down to decrease the inflow. This example illustrates the concept of ``negative feedback'': the control action compensates for the departure from the desired water level, irrespective of what causes this departure. We see that the control system self-regulates: regardless of the actions of the adversary (letting more or less water out of the sink), the controller always takes the action that brings the water level back to the reference height.

Negative feedback loops appear in many areas such as biology (e.g., regulating cell cycle), environment (e.g., stabilizing Earth climate), and engineering (e.g., James Watt's 1788 centrifugal mechanism that controlled the speed of his steam engine).
The concept of negative feedback was mathematically formalized by \cite{maxwell1868governors}, and \cite{wiener1948cybernetics}. 


In this paper, we propose a novel adversarial training approach based on control theory. Specifically, {\it we develop a new neural network architecture, named Feedback Neural Networks, that incorporates feedback control.} The original neural network is extended with a controller that feeds output discrepancy information back to the input of the original network. The controller is itself a neural network, that is trained using regular and adversarial data such as to stabilize the system outputs. {\it The novel adversarial training approach based on the feedback control architecture is called Feedback Looped Adversarial Training (FLAT).}

The incorporation of negative feedback in the new architecture leads to a structure that is capable to self-correct for output errors regardless of the perturbations applied to the inputs, and help the original network in making correct predictions.

The structure of the paper is as follows:
\begin{itemize}
    \item The Background section reviews the formulation of adversarial training, as well as fundamentals of control theory.
    \item The Methodology section develops the proposed adversarial training method based on feedback control.
    \item Empirical evidence of the effectiveness of our proposed method is provided in the Experimental Results section.
    \item Final remarks and directions of future research are given in the Conclusions section.
\end{itemize}

\section{Background}
\label{sec:background}

\subsection{Adversarial Training Formulation}

We consider the task of classification using a neural network (NN) model \cite{lawrence1993introduction}
\begin{align}
\label{eqn:NN}
\mathbf{Y} = f(\mathbf{X}, \boldsymbol{\theta}),
\end{align}
where $\mathbf{X}$ is some collection of inputs, and $\boldsymbol{\theta}$ is the set of trainable parameters of the NN model. The task consists of classifying the inputs into one of $\mathbf{C}$ classes/labels accurately considering ground truth labels $\mathbf{Y}\in \{1, 2, \ldots, \mathbf{C}\}$. 

We also consider well-known adversarial attacks that produce inputs $\mathbf{X} + \Delta \mathbf{X}$, where $\Delta \mathbf{X}$ is some modification to the original input that is imperceptible to the human eye. Usually, a predefined range of small perturbations  is considered, e.g., $\Vert \Delta \mathbf{X} \Vert_\infty \le \varepsilon \in \mathbb{R}$. The goal of adversarial training is to find parameter $\boldsymbol{\theta}$ values such that the NN model \eqref{eqn:NN} is robust against these kinds of attacks, i.e., the small perturbation does not change the classification: $f(\mathbf{X} +  \Delta \mathbf{X}, \boldsymbol{\theta}) = f(\mathbf{X}, \boldsymbol{\theta})$ for any  
sufficiently small $\Vert \Delta \mathbf{X} \Vert$.

One popular formulation of adversarial training involves solving a min-max optimization problem \cite{goodfellow2014explaining}. Specifically, for some loss function $\mathcal{L}(\mathbf{Y}, f(\mathbf{X}, \boldsymbol{\theta}))$, the training optimization problem is formulated as follows:

\begin{align}
\label{eqn:min-max-optimization}
    \argmin_{\boldsymbol{\theta}} \argmax_{\Delta \mathbf{X} \,:\, \Vert \Delta \mathbf{X} \Vert_\infty \le \varepsilon} \mathcal{L}(\mathbf{Y}, f(\mathbf{X} + \Delta \mathbf{X}, \boldsymbol{\theta})).
\end{align}

\subsection{Feedback Control}

In control theory \cite{lee1967foundations, doyle2013feedback, woolf2009chemical}, a system is stabilized by passing the output of the system that is measured using a sensor (or the error between the output and a reference signal) through a controller. The controller manipulates the input and computes a control action that is fed back  into the system. For example, the scaled output error is subtracted from the reference input. The output information is passed backward to the input, whence the name {\it feedback} control; and the subtraction ensures that the sign of the change compensates for (acts against) the output errors, whence the name {\it negative feedback}. This mechanism helps stabilize system response in the presence of noise or perturbations in the input; we note that stabilization happens regardless of the perturbation, since whatever $\Delta \mathbf{X}$ change is applied to the inputs, the negative feedback  control action always nudges the output back toward the correct (unperturbed) value. This architecture is known as a closed loop controller or a feedback controller. 

\begin{figure*}[htpb]
    \centering
    \includegraphics[scale=0.8]{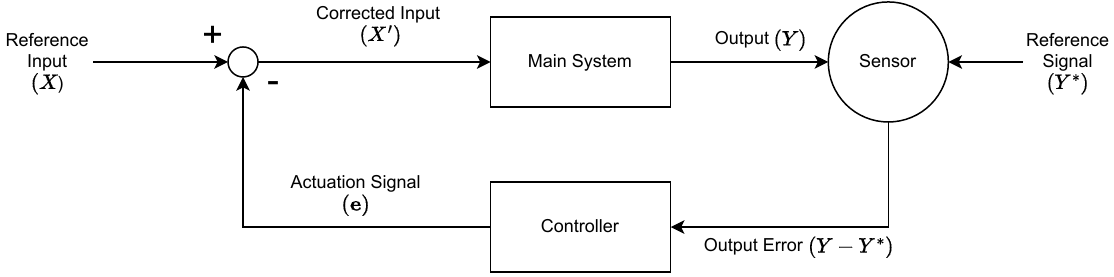}
    \caption{Block diagram of a system with a negative feedback loop for error correction.}
    \label{fig:control-theory}
\end{figure*}

The idea is illustrated in Figure \ref{fig:control-theory}. An input $(\mathbf{X})$ is passed to the main system and then the state/output $(\mathbf{Y} = f(\mathbf{X}, \boldsymbol{\theta}))$  of the main system is read by the sensor. The output error, i.e., the difference between the output $(\mathbf{Y})$ and a reference signal $(\mathbf{Y}^\ast)$ is computed. The sensor passes the output error $(\mathbf{Y}-\mathbf{Y}^\ast)$ to a controller system that computes an actuation signal $(\mathbf{e})$ that helps in stabilizing the system. The job of the controller is to predict the correction $(\mathbf{e})$ of the input $X$, such that the corrected input $(\mathbf{X}' = \mathbf{X}-\mathbf{e})$ passed through the main system leads to the reference output ($\mathbf{Y}$). Passing the information through the system, the controller, and then the system again, can be done several times until equilibrium is reached. By definition, the feedback in this system is known as negative feedback. In negative feedback, an increase/decrease in some variable triggers an opposite change (decrease/increase) in another variable. In the system in Figure \ref{fig:control-theory}, the controller takes corrective action on the input, which forces the output towards the desired output. This makes the system oscillate around a stable state. In other words, if the output increases, the input signal will be decreased to compensate for that and vice versa.

\paragraph{Example.}
Consider a linear main system described by the symmetric matrix 
\[
\mathbf{A} = \varepsilon^{-1}\,v_1\, v_1^T + \sum_{i=2}^n v_i\, v_i^T \in \mathbb{R}^{n \times n}, \quad \Vert v_i \Vert =1, 
\] 
Here $(\lambda_1,v_1)$ be the dominant eigenpair with $\lambda_1 = \varepsilon^{-1}$, $\varepsilon \ll 1$; all other eigenpairs $(\lambda_i,v_i)$ have $\lambda_i  = 1$, $i\ge 2$, and all eigenvectors are orthonormal. 
Let the reference input be $\mathbf{X}=0$, with the reference output  $\mathbf{Y}=0$.
A small input perturbation $\Delta\mathbf{X} = \varepsilon\, v_1$ with $\Vert \Delta\mathbf{X} \Vert = \varepsilon$ leads to an output change $\Delta \mathbf{y} = \mathbf{A}\cdot\Delta \mathbf{X} = v_1$ of size $\Vert \Delta \mathbf{y} \Vert = 1$.

Consider now a linear controller described by the matrix $\mathbf{K} \in \mathbb{R}^{n \times n}$. For the closed loop system in Figure \ref{fig:control-theory} the input-output relation is:
\begin{equation}
\label{eqn:linear-feedback}
\mathbf{y} 
= (\mathbf{I}-\mathbf{A}\,\mathbf{K})^{-1}\,\mathbf{A}\cdot\mathbf{x}.
\end{equation}
The size of the output perturbation \eqref{eqn:linear-feedback} can be made small by suitably choosing the control matrix. For example, for $\mathbf{K} = \kappa\, v_1\, v_1^T$  the output perturbation has the size $\Vert \Delta \mathbf{y} \Vert = |1 / (1 - \kappa\,\varepsilon^{-1})|$, which is $\mathcal{O}(\varepsilon)$ for $\kappa = \mathcal{O}(1)$.

The exact input-output equation \eqref{eqn:linear-feedback} involves the inverse of the matrix $(\mathbf{I}-\mathbf{K})$. To avoid this, one can iteratively pass the input information through the main system, then feed back through the controller and form the corrected input; the next iteration repeats the process with the corrected input:
\begin{eqnarray}
\nonumber
&& \mathbf{y}^{(i)} = \mathbf{A}\,\mathbf{x}^{(i)}, \\
\label{eqn:linear-feedback-iterative}
&&\mathbf{x}^{(i+1)} = \mathbf{x}^{(i)} - \mathbf{K}\,\mathbf{y}^{(i)} = (\mathbf{I}-\mathbf{K}\mathbf{A})\mathbf{x}^{(i)}, \\
\nonumber
&& \mathbf{y}^{(P)} = \mathbf{A}\,\mathbf{x}^{(P)} =  \mathbf{A}\,(\mathbf{I}-\mathbf{K}\mathbf{A})^P\,\mathbf{x}^{(0)}.
\end{eqnarray}
With $\mathbf{K} = \kappa\, v_1\, v_1^T$ we have $\Vert \mathbf{y}^{(P)} \Vert = |1 - \kappa/\varepsilon|^P$, and for 
$\kappa \approx \varepsilon$ with $|1 - \kappa/\varepsilon| < 1$ the output perturbation decreases progressively with increasing the number of iterations  $P$. For $\kappa \approx \varepsilon\,(1-\varepsilon)$ after one iteration we have  $\Vert \mathbf{y}^{(1)} \Vert = \mathcal{O}(\varepsilon)$.
We see that a suitable control gain $\kappa$ (in general, a suitable control feedback matrix $\mathbf{K}$) is chosen differently for iterated feedback \eqref{eqn:linear-feedback-iterative} than for the exact feedback \eqref{eqn:linear-feedback}.

\section{Methodology}
\label{sec:method}

\subsection{Feedback Neural Networks}

We apply the negative feedback control principle discussed in the Background section to robustify neural networks against adversarial attacks. Consider the general control diagram illustrated in Figure \ref{fig:control-theory}. In the proposed approach the main system represents the original NN model we want to robustify, and the sensor represents its output layer. The controller in Figure \ref{fig:control-theory}  is implemented by a trainable control NN that will help stabilize the original NN model. 

\begin{figure*}[!tb]
    \centering
    \includegraphics[scale=0.9]{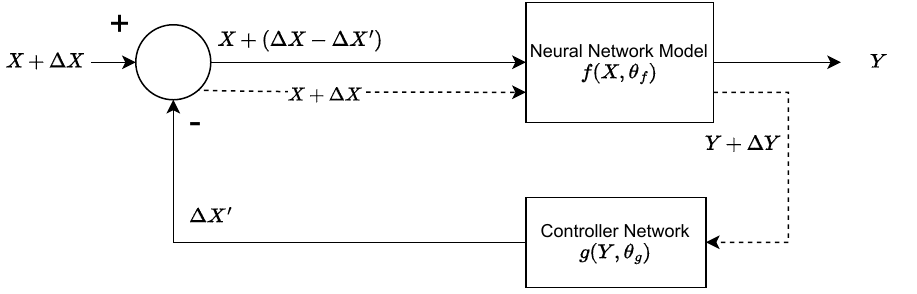}
    \caption{Architecture of the Feedback NN model. The original NN to be stabilized $f(\mathbf{X},\boldsymbol{\theta}_f)$ is added a negative feedback loop. The controller of this loop is another trainable NN $g(\mathbf{Y},\boldsymbol{\theta}_g)$.}
    \label{fig:nn-defense}
\end{figure*}

The resulting feedback NN architecture is depicted in Figure \ref{fig:nn-defense}. Consider a reference input $\mathbf{X}$, and the corresponding reference NN output $\mathbf{Y} = f(\mathbf{X}, \mathbf{\boldsymbol{\theta}}_f)$. Consider next a perturbed input $\mathbf{X} + \Delta \mathbf{X}$, which is passed through the NN model, to obtain a predicted output  $\mathbf{Y} + \Delta \mathbf{Y} = f(\mathbf{X}+ \Delta \mathbf{X}, \mathbf{\boldsymbol{\theta}}_f)$. The predicted output  is then passed to the controller NN model, which produces the feedback perturbation ($\Delta \mathbf{X}' = g(\mathbf{Y}+ \Delta \mathbf{Y}, \mathbf{\boldsymbol{\theta}}_g)$ (dashed line in Figure \ref{fig:nn-defense}). The feedback perturbation ($\Delta \mathbf{X}'$) is added to the original input, and seeks to undo the effect of the unknown original data perturbation. Specifically, the corrected input $\widetilde{\mathbf{X}} = \mathbf{X} + \Delta \mathbf{X} - \Delta \mathbf{X}'$ is such that, when propagated through the NN, it approximately recovers the reference output: $\mathbf{Y}  = f(\widetilde{\mathbf{X}}, \mathbf{\boldsymbol{\theta}}_f)$. (A sufficient, but not necessary, condition is that $\Delta \mathbf{X} - \Delta \mathbf{X}' \approx 0$, i.e., the controller is trained to predict the unknown input perturbation, which is then subtracted from the adversarially corrupted input.) 

Determining the optimal architecture of a controller network for a given main NN is a separate research problem, beyond the scope of this paper, and will be addressed in future work. For simplicity, here we employ a standard feed-forward architecture for the controller neural networks. 

\subsection{Implementation of  Feedback Neural Networks}

\begin{figure*}
    \centering
    \includegraphics[scale=1.1]{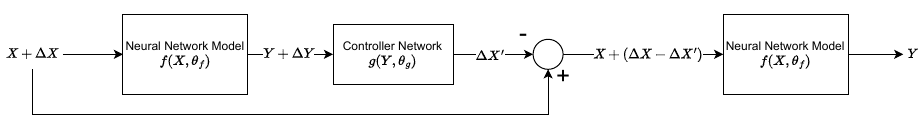}
    \caption{Unrolled Architecture of the Feedback NN model. The ``residual network'' part (first two blocks from the left) can be repeated $P$ times to implement $P$ unrolled iterations.}
    \label{fig:nn-defense-unrolled}
\end{figure*}

Consider the generic feedback NN shown in Figure \ref{fig:nn-defense}.
The proposed implementation uses an unrolled version of this architecture, and the unrolled architecture is depicted in Figure \ref{fig:nn-defense-unrolled}. We describe this implementation in PyTorch \cite{paszke2017automatic}. We define a new feedback network that takes as input the output of a predefined model that we want to robustify. So, the predefined model and the feedback network are connected and we consider this whole network as a new model.
During the forward pass of the newly defined model, the perturbed input $\mathbf{X} + \Delta \mathbf{X}$ is passed to the predefined model $f(\mathbf{X}, \mathbf{\boldsymbol{\theta}}_f)$, to produce the output $\mathbf{Y} + \Delta \mathbf{Y}$ (perturbed from the reference due to the presence of input perturbation). This output is passed to the controller/feedback network $g(\mathbf{X}, \mathbf{\boldsymbol{\theta}}_g)$, that produces the correction input signal $\Delta \mathbf{X}' \approx \Delta \mathbf{X}$. The correction signal is subtracted from the given input to obtain $\mathbf{X} + \Delta \mathbf{X} - \Delta \mathbf{X}'$. The corrected input is passed again to the predefined model $f(\mathbf{X}, \mathbf{\boldsymbol{\theta}}_f)$ again to produce an output that should be closer to the desired one. This cycle can be iterated several times, i.e, pass the corrected output again through the controller network to produce an even better result. During our numerical experiments, we observed that passing information through the controller network multiple times did not significantly improve the robust accuracy. 

\subsection{Adversarial Training of Feedback Neural Networks}

We slightly modify the adversarial training procedure described in \cite{madry2017towards} and adapt it to feedback NNs. The training algorithm is given in Algorithm \ref{alg:adversarial-training}. In the algorithm, we consider the original NN model together with the control feedback model as an aggregate NN model $\mathbf{F}$ with learnable parameters $\boldsymbol{\theta} = \{\boldsymbol{\theta}_f, \boldsymbol{\theta}_g\}$. The $\Pi_{\varepsilon}$ symbol in the algorithm in line number 6 denotes the projection onto the ball of interest; here we use the $L_\infty$ norm for projection. For creating the perturbed points, we consider the PGD attack \cite{madry2017towards}. We name this procedure Feedback Looped Adversarial Training (FLAT).

\begin{algorithm}[tb]
\caption{Feedback Looped Adversarial Training (FLAT).}
\label{alg:adversarial-training}
\textbf{Input}: A feedback NN model $\mathbf{F}$ (including both the original and the control NNs, as in Figure \ref{fig:nn-defense-unrolled}) with learnable parameters $\boldsymbol{\theta}$; loss function $\mathcal{L}$; input data $\mathbf{X}$; ground truth $\mathbf{Y}$; number of epochs/iterations $N$; number of adversarial attack steps $K$; magnitude of maximum allowed perturbation $\varepsilon$; perturbation step size $\kappa$; learning rate $\tau$. 
\begin{algorithmic}[1]
    \STATE Initialize $\boldsymbol{\theta}$
    \FOR{$i\leftarrow 1, 2, \ldots, N$}
        \STATE Sample batches $(x, y)$ from input data $\mathbf{X}$ and ground truth $\mathbf{Y}$
        \STATE Update parameters: $\boldsymbol{\theta} = \boldsymbol{\theta} - \tau\nabla_{\boldsymbol{\theta}} \mathcal{L} (y, \mathbf{F}(x, \boldsymbol{\theta}))$
        \FOR{$k \leftarrow 1, 2, \ldots, K$}
            \STATE $x'_k \leftarrow \Pi_{\varepsilon}(\kappa \sign (x'_{k-1} + \nabla_{x'_{k-1}} \mathcal{L} (y, \mathbf{F}(x'_{k-1}, \boldsymbol{\theta})))$    
        \ENDFOR
        \STATE Update parameters: $\boldsymbol{\theta} = \boldsymbol{\theta} - \tau\nabla_{\boldsymbol{\theta}} \mathcal{L} (y, \mathbf{F}(x'_k, \boldsymbol{\theta}))$
    \ENDFOR
    \STATE \textbf{return} Trained NN model $\mathbf{F}$ with updated parameters $\boldsymbol{\theta}$
\end{algorithmic}
\end{algorithm}

Normally, in adversarial training, the parameter updates are done only by considering the perturbed data points during calculation of loss values and its respective gradients. But, in our preliminary experiments, we have noticed that only doing this results in the model performing well only on the perturbed data points but not on the original data points. We speculate that without the original data as reference, the feedback network is not able to differentiate between the perturbed data and the original data. That is why, in the algorithm, we consider both the original data points and the perturbed data points.

\section{Experimental Results}

\label{sec:experiment}


\begin{table*}[!t]
    \centering
    
    \begin{tabular}{l|cccccc}
    Method & NAT & FGSM & PGD$^{20}$ & PGD$^{100}$ & MIM & CW\\
    \hline
    \hline
         AT &  $84.32 \pm 0.23$ & $51.70\pm0.12$ & $48.29\pm0.11$ & $48.12\pm0.13$ & $47.95\pm0.04$ & $49.57\pm0.15$\\
         TRADES & $83.91 \pm 0.33$ & $62.82\pm0.13$ & $54.25\pm0.11$ & $52.21\pm0.09$ & $55.65\pm0.10$ & $52.22\pm0.05$\\
         FAT & $\mathbf{87.72 \pm 0.14}$ & $58.10\pm0.25$ & $46.69\pm0.31$ & $46.81\pm0.30$ & $47.03\pm0.17$ & $49.66\pm0.38$\\
         MART & $83.12 \pm 0.23$ & $65.65\pm0.23$ & $55.43\pm0.16$ &  $53.46\pm0.24$ & $57.06\pm0.20$ & $51.45\pm0.29$\\
         GAIRAT & $83.40 \pm 0.21$ & $63.00\pm0.18$ & $54.76\pm0.42$ & $54.81\pm0.63$ & $53.57\pm0.31$ & $38.71\pm0.26$\\
         SEAT & $83.70 \pm 0.13$ & $54.77 \pm 0.05$ & $56.02\pm0.11$ & $55.97\pm0.07$ & $57.13\pm0.12$ & $54.38\pm0.10$\\
         \hline
         \hline
         \textbf{FLAT} & $84.74\pm0.21$ & $72.25\pm0.35$ & $59.56\pm0.40$ & $59.46\pm0.31$ & $60.27\pm0.33$ & $59.65\pm0.18$\\
         \textbf{FLAT (feat.)} & $84.73\pm0.15$ & $\mathbf{72.43\pm0.36}$ & $\mathbf{59.79\pm0.37}$ & $\mathbf{59.63\pm0.22}$ & $\mathbf{60.30\pm0.27}$ & $\mathbf{59.74\pm0.23}$\\
         
    \end{tabular}
    \caption{Accuracy results under different attacks against ResNet-18 model adversarially trained on CIFAR-10 dataset. (NAT represents no attack.) Entries represent mean accuracies   reported in ($\%$ of the reference value) along with the standard deviation over multiple runs. The highest accuracy for each attack method is in bold font.}
    \label{tab:resnet18-cifar10-results}
\end{table*}

\begin{table*}[!t]
    \centering
    
    \begin{tabular}{l|cccccc}
    Method & NAT & FGSM & PGD$^{20}$ & PGD$^{100}$ & MIM & CW\\
    \hline
    \hline
         AT &  $87.32\pm0.21$ & $56.10\pm0.25$ & $49.01\pm0.33$ & $48.83\pm0.27$ & $48.25\pm0.17$ & $52.80\pm0.25$\\
         TRADES & $85.11\pm0.77$ & $61.06\pm0.33$ & $54.58\pm0.49$ & $54.82\pm0.38$ & $55.67\pm0.31$ & $54.91\pm0.21$\\
         FAT & $89.65\pm0.04$ & $65.52\pm0.15$ & $48.74\pm0.23$ & $48.69\pm0.18$ & $48.24\pm0.16$ & $52.11\pm0.71$\\
         MART & $84.26\pm0.28$ & $67.51\pm0.11$ & $54.11\pm0.58$ & $54.13\pm0.30$ & $55.20\pm0.22$ & $53.41\pm0.17$\\
         GAIRAT & $85.92\pm0.69$ & $65.00\pm0.31$ & $58.51\pm0.42$ & $58.48\pm0.34$ & $58.37\pm0.27$ & $44.31\pm0.22$\\
         SEAT & $86.44\pm0.12$ & $59.23\pm0.03$ & $59.84\pm0.20$ & $59.80\pm0.16$ & $60.87\pm0.10$ & $58.95\pm0.34$\\
         \hline
         \hline
         \textbf{FLAT} & $93.61\pm0.29$ & $83.85\pm0.41$ & $73.10\pm0.30$ & $72.77\pm0.33$ & $74.59\pm0.26$ & $73.02\pm0.26$\\
         \textbf{FLAT (feat.)} & $\mathbf{93.71\pm0.24}$ & $\mathbf{84.09\pm0.18}$ & $\mathbf{73.35\pm0.38}$ & $\mathbf{73.13\pm0.32}$ & $\mathbf{74.77\pm0.21}$ & $\mathbf{73.43\pm0.35}$\\
         
    \end{tabular}
    \caption{Accuracy results under different attacks against WRN-32-10 model adversarially trained on CIFAR-10 dataset. (NAT represents no attack.) Entries represent mean accuracies   reported in ($\%$ of the reference value) along with the standard deviation over multiple runs. The highest accuracy for each attack method is in bold font.}
    \label{tab:wrn3210-cifar10-results}
\end{table*}

\begin{table*}[!t]
    \centering
    
    \begin{tabular}{l|cccccc}
    Method & NAT & FGSM & PGD$^{20}$ & PGD$^{100}$ & MIM & CW\\
    \hline
    \hline
         AT &  $54.06 \pm 0.35$ & $39.89\pm0.18$ & $29.31\pm0.30$ & $29.99\pm0.12$ & $30.82\pm0.41$ & $29.26\pm0.51$\\
         TRADES & $59.93 \pm 0.46$ & $35.45 \pm 0.32$ & $29.90\pm0.41$ & $29.88\pm0.11$ & $29.55\pm0.25$ & $26.14\pm0.21$\\
         FAT & $\mathbf{61.50\pm0.23}$ & $32.17\pm0.31$ & $20.00\pm0.24$ & $19.00\pm0.15$ & $22.32\pm0.26$ & $21.34\pm0.13$\\
         MART & $57.24 \pm 0.64$ & $39.21\pm 0.10$ & $30.62\pm0.17$ &  $30.62\pm0.17$ & $30.83\pm0.28$ & $26.30\pm0.29$\\
         GAIRAT & $56.00\pm0.11$ & $30.10\pm0.06$ & $19.05\pm0.24$ & $19.15\pm0.22$ & $21.00\pm0.32$ & $18.50\pm0.27$\\
         SEAT & $56.28 \pm 0.33$ & $-$ & $\mathbf{32.15\pm0.17}$ & $\mathbf{32.12\pm0.26}$ & $\mathbf{32.62\pm0.15}$ & $\mathbf{29.68\pm0.26}$\\
         \hline
         \hline
         \textbf{FLAT} & $54.99\pm0.49$ & $\mathbf{40.38\pm0.25}$ & $30.12\pm0.25$ & $30.13\pm0.43$ & $30.95\pm0.30$ & $29.54\pm0.27$\\
         \textbf{FLAT (feat.)} & $54.68\pm0.31$ & $40.16\pm0.31$ & $29.89\pm0.21$ & $29.84\pm0.28$ & $30.70\pm0.28$ & $29.31\pm0.21$\\
         
    \end{tabular}
    \caption{Accuracy results under different attacks against ResNet-18 model adversarially trained on CIFAR-100 dataset. Entries represent mean accuracies   reported in ($\%$ of the reference value) along with the standard deviation over multiple runs. The highest accuracy for each attack method is in bold font.}
    \label{tab:resnet18-cifar100-results}
\end{table*}
\begin{table*}[!tb]
    \centering
    
    \begin{tabular}{l|cccccc}
    Method & NAT & FGSM & PGD$^{20}$ & PGD$^{100}$ & MIM & CW\\
    \hline
    \hline
         AT &  $\mathbf{71.98\pm0.33}$ & $\mathbf{55.38\pm0.28}$ & $\mathbf{45.74\pm0.12}$ & $45.60\pm0.17$ & $\mathbf{46.94\pm0.13}$ & $45.15\pm0.29$\\
         TRADES & $57.29\pm0.41$ & $36.25\pm0.12$ & $27.45\pm0.31$ & $27.21\pm0.14$ & $29.87\pm0.21$ & $27.50\pm0.37$\\
         FAT & $66.74\pm0.34$ & $38.25\pm0.21$ & $24.50\pm0.10$ & $24.00\pm0.16$ & $27.21\pm0.33$ & $25.45\pm0.23$\\
         MART & $58.96\pm0.37$ & $41.83\pm0.20$ & $34.43\pm0.35$ & $34.38\pm0.21$ & $35.64\pm0.37$ & $31.32\pm0.16$\\
         GAIRAT & $61.50\pm0.24$ & $35.15\pm0.09$ & $24.32\pm0.41$ & $24.00\pm0.46$ & $26.75\pm0.32$ & $24.35\pm0.26$\\
         SEAT & $62.34\pm0.05$ & $34.74\pm0.15$ & $35.10\pm0.43$ & $35.15\pm0.35$ & $35.70\pm0.23$ & $33.03\pm0.17$\\
         \hline
         \hline
         \textbf{FLAT} & $71.87\pm0.28$ & $55.01\pm0.44$ & $45.52\pm0.27$ & $45.28\pm0.34$ & $46.52\pm0.39$ & $45.06\pm0.44$\\
         \textbf{FLAT (feat.)} & $71.82\pm0.19$ & $55.04\pm0.19$ & $45.45\pm0.22$ & $\mathbf{45.63\pm0.24}$ & $46.38\pm0.15$ & $\mathbf{45.15\pm0.20}$\\
         
    \end{tabular}
    \caption{Accuracy results under different attacks against WRN-32-10 model adversarially trained on CIFAR-100 dataset. Entries represent mean accuracies   reported in ($\%$ of the reference value) along with the standard deviation over multiple runs. The highest accuracy for each attack method is in bold font. All methods perform equally well.}
    \label{tab:wrn3210-cifar100-results}
\end{table*}


\subsection{Experimental Setup}

We perform our experiments using ResNet18 \cite{he2016deep} and WRN-32-10 \cite{zagoruyko2016wide} convolutional networks. For training and testing purposes, we use the CIFAR-10 and CIFAR-100 dataset \cite{krizhevsky2009learning} that consists of images of dimension $32\times32\times3$ of 10 categories and 100 categories respectively. We train ResNet18 for 120 epochs with an initial learning rate of 0.01 and batch size 128. We optimize ResNet18 using SGD with 0.9 momentum and weight decay factor of $3.5\times10^{-3}$. For WRN-32-10, the number of epochs and batch size remains the same. We change the initial learning rate to 0.1 and weight decay factor to $7\times10^{-4}$. For both models, we maintain the initial learning rate ($lr$) until epoch 40, then linearly reduce it to $lr/10$ until epoch 80, and then linearly reduce it again to $lr/100$ until epoch 120. The magnitude of maximum perturbation  is $\varepsilon = 8/255$ with  perturbation step size $\kappa = 2/255$. The number of steps for the PGD attack during adversarial training is 10. We train the model for 10 random runs started from 10 random initializations. We choose to use the aforementioned models and datasets for two reasons: i) the relevant gradient computations can be done in a reasonable amount of time, and ii) these datasets and models have been used extensively in the literature for comparison purposes, and provide an excellent benchmark.

\subsection{Controller Network}

For both ResNet18 and WRN-32-10, we use a simple controller network architecture with fully connected layers.  The input to the controller network is chosen in two different ways.
\begin{enumerate}
\item In the first version (simple FLAT) the controller input comes from the final prediction vector produced by the main model. The controller network has 3 layers of 512, 1024 and 3072 neurons, respectively. We reshape the output from the 3rd layer of the controller network to the dimension of the input image so that we can do error correction on the original input image. 
\item In the second version (FLAT (feat.)) the controller input consists of the features produced by the last ResNet block of both models, concatenated with the prediction vector of the main model. In the second case the controller network has 3 fully connected layers of 1024, 1024 and 3072 neurons respectively.
\end{enumerate}

\subsection{Attack Methods and Comparison Against State-of-the-Art Methods}

We evaluate the robustness of the trained models against several attacks including FGSM \cite{goodfellow2014explaining}, PGD attack \cite{madry2017towards}, MIM \cite{dong2018boosting} and CW \cite{carlini2017towards}. We also evaluate the performance of the trained models on the natural datasets with no perturbations (NAT). 

To evaluate the effectiveness of our method, we compare it against the following state-of-the-art algorithms: standard AT \cite{madry2017towards}, TRADES \cite{zhang2019theoretically}, MART \cite{wang2019improving}, FAT \cite{zhang2020attacks}, GAIRAT \cite{zhang2020geometry} and SEAT \cite{wang2022self}.

\subsection{Robustness Evaluation under Adversarial Attacks}

We now present the numerical results of our experiments.
In the literature, classification accuracy is the preferred metric used to measure the robustness of a classification model under adversarial attacks. For each experiment we perform 10 different runs, and report the average accuracy results and the corresponding standard deviations in Tables \ref{tab:resnet18-cifar10-results} and \ref{tab:wrn3210-cifar10-results}. 

The results for ResNet18 on CIFAR-10 are reported in Table \ref{tab:resnet18-cifar10-results}. Our FLAT method outperforms previous methods. It shows and improvement of $\sim 6\%$ compared to PGD$^{20}$, PGD$^{100}$ and MIM. It also shows an improvement of $\sim10\%$ over FGSM and CW. Only FAT performs better than our method, and only when no attacks are performed. 

Results on WRN-32-10 on CIFAR-10 are reported in Table \ref{tab:wrn3210-cifar10-results}. The performance improvement achieved by our method is even greater for this model. Our method improves accuracy both on natural data and perturbed data. Against all attacks, our method managed to improve the robust accuracy by $\sim 20-24\%$ compared to the previous methods.

The best performances are achieved by our method variation, where we consider some features from the models along with the predictions. The variation where we do not consider the features also outperforms the previous methods. However, the improvement when we consider the features is minimal.

From the Tables \ref{tab:resnet18-cifar10-results} and \ref{tab:wrn3210-cifar10-results}, we can see that there is a significant improvement on robust accuracies achieved by the previous methods. 

Tables \ref{tab:resnet18-cifar100-results} and \ref{tab:wrn3210-cifar100-results} tabulates the performance of ResNet18 and WRN-32-10 on CIFAR-100 respectively. On CIFAR-100, the performance of ResNet18 is below par even with our method. ResNet18 is not a good enough model to perform well on CIFAR-100 even without any perturbations. In the case of WRN-32-10, the performance is overall equally well as the standard AT. But, we think future work on how the feedback network should be in a given scenario will help us improve the performance more.

\subsection{Implementations and Resources}

We used PyTorch \cite{paszke2017automatic} to implement all experiments. We conducted all our experiments on a machine with AMD EPYC 7742 CPU @ 2.2GHz with 2 TB RAM and NVIDIA A100 GPU with 80 GB memory. The datasets are publicly available online and also through PyTorch function calls. We will add the GitHub link to our repository after the anonymous review process.


\section{Conclusions and Future Developments}
\label{sec:conclusion}

We propose a new method for neural network robustification inspired from control theory. To this end, we develop a new ``Feedback NN'' architecture that adds a negative feedback loop to the original NN for error correction. The controller, which map errors in the outputs back to corrections of the inputs, is itself a trainable neural network. The extended system (main model plus controller) is adversarially trained on perturbed data sets, which leads to the novel \textit{Feedback looped adversarial training (FLAT).} Experimental results on standard tests show that our FLAT method is more effective than the state-of-the-arts against various attacks. 

Owing to the power of the negative loop concept, relatively simple and small controller architectures may suffice for robustifying much larger main NN models. Here we use simple 3-layers feed forward controllers. In the future, we plan to study additional controller network architectures, and seek to design optimal controllers for the robustification of given NN models. 

In this work the information used in the feedback loop (i.e., the inputs to the controller) consists of errors in the main NN model output, and the features from the last output layer. Future work will study the use of features from other layers of the original neural network to improve the resulting robustness of the Feedback NN.

\bibliography{aaai24}

\end{document}